# Energy Use of AI Inference, Efficiency Pathways, and Test-Time Scaling

Felipe Oviedo*, Fiodar Kazhamiaka, Esha Choukse, Allen Kim, Amy Luers, Melanie Nakagawa, Ricardo Bianchini, Juan M. Lavista Ferres

**Microsoft, 1 Microsoft Way, Redmond WA 98052, USA**

*Lead Contact: felipe.oviedo@microsoft.com

## SUMMARY

As AI inference scales to billions of queries, estimates of per-query energy use are increasingly important for capacity planning, efficiency interventions, and policy. Yet many public estimates assume non-production settings, leading to systematic overestimation. We introduce a bottom-up framework estimating inference energy from token throughput, node power, and overhead under large-scale deployment assumptions. For frontier-scale models (>200B parameters) on H100 nodes, we estimate a median energy of 0.31 Wh/query (IQR 0.16–0.60), indicating widely cited estimates are overstated by 4–20×. In test-time scaling scenarios 15× longer than typical queries, the median energy rises 13× to 3.91 Wh (IQR 2.15–7.05). Across models, serving systems, and hardware, we estimate 8–20× line-of-sight energy reductions. At datacenter scale, serving 1 billion queries/day requires 0.7 GWh; if 10% are long queries, demand rises to 1.7 GWh/day. With efficiency interventions, it falls to 0.8 GWh/day, mitigating the energy impact of test-time scaling.

## 1 INTRODUCTION

LLM (Large Language Model) inference is one of the fastest-growing sources of computing demand, with energy use already approaching that of web search and other mature digital services.[1–3] As a result, many stakeholders are seeking reliable figures on how much energy is consumed during inference at a granular level (e.g., energy per textual query) to provide a firm basis for energy projections, compare serving approaches, and relate efficiency considerations to engineering and product trade-offs. Despite this need, public reports of "energy per query" differ by more than an order of magnitude.[2,4]

Two issues drive this discrepancy. First, measurement boundaries vary: some account only for GPU draw, others include CPUs or application overhead. Second, many benchmarks fail to capture realistic production conditions such as batching, concurrency, and steady-state serving.[5–7] Model-based estimates are also highly sensitive to assumptions about hardware specifications, parameter counts, and token lengths.[8–10] Direct measurements, though more precise, are often tied to non-production models or small-batch tests.[6] These settings tend to overestimate energy use relative to large, high-concurrency deployments, which handle most AI inference workloads.

An added challenge to characterizing inference energy is that per-query figures cannot be interpreted in isolation. The energy used in inference also depends on how energy demand scales when billions of queries are served, and how interventions at the model, serving software, and hardware layers can moderate that growth. Recent model provider disclosures offer a more realistic baseline[2,11] by measuring large-scale systems but remain limited by confidentiality. Typically, these reports only include a single "representative" or "average query", masking how energy use changes with query length distributions and model and serving configurations. This limitation is significant: test-time scaling (reasoning models) and agentic workflows are increasingly routine, and their long outputs can drive a disproportionate share of aggregate energy demand. Without this system context, averages give a misleading picture of energy use in inference. Furthermore, such metrics fail to provide guidance on where future efficiency efforts should focus.

With these limitations in mind, we propose a bottom-up estimation approach to fill the gaps. Our framework aligns with production-scale disclosures but adds two contributions: (i) a decomposition of the interacting determinants of energy per query—model, serving platform, and hardware—to provide insight into practical efficiency interventions; and (ii) explicit modeling of token-length distributions, showing how long inference increases energy per query.

## 2 METHODS

Throughout this work, we use the most common definition of a "query":[2,8,9] a single text prompt–completion event, consisting of a single input sequence and the LLM-generated output sequence produced in response under steady-state serving conditions. To estimate the per-query energy use, we focus on at-scale deployments where nodes are saturated through batching, high concurrency, and optimized serving. LLM inference has two stages: prefill, where the input sequence is processed once in parallel, and decoding, where output tokens are generated step by step. Because decoding cannot be parallelized across tokens, it often dominates energy use in most real-world workloads—a trend amplified in reasoning and agentic workflows,[12] or when multiple sequences are generated per request (best-of-n-sampling).

Our estimation framework is designed around these conditions. We propose a bottom-up Monte Carlo simulation that captures these interacting factors for various open-source models. For a fixed GPU node, the per-query energy $E_{\text{query}}$ (Wh/query) is estimated as:

$$E_{\text{query}} = \frac{\text{PUE}}{3.6}\left(\frac{P_{\text{node}} \cdot L_{\text{eff}}}{TPS(L_{\text{in}}, L_{\text{out}})}\right) \cdot \frac{1}{\propto} \quad \textbf{(Equation 1)}$$

Here, $P_{\text{node}}$ is the steady-state power draw per GPU node during inference (kW). For our analysis, we assume models are hosted on 8xH100 GPUs at FP8 precision, as in the NVIDIA DGX H100 architecture,[13] or 10xH100 (in the case of DeepSeek-R1), with a maximum power draw $P_{\max}$ of 10.2 kW or 12.5 kW (linearly scaled), respectively. For highly utilized nodes, average power utilization has been reported around 70% during inference.[12,4] Based on these studies, we model $P_{\text{node}}$ as a log-normal distribution centered at $0.7 \cdot P_{\max}$ with P5–95 range of $0.4 \cdot P_{\max}$– $0.9 \cdot P_{\max}$. The aggregated effect of idle nodes and overhead is explored below. PUE is the power usage effectiveness of the AI data center, modeled as a log-normal distribution with P5–P95 range of 1.05–1.40, consistent with hyperscaler public reports[14,15]. The constant 3.6 converts kW to Wh. The efficiency factor $\propto$, defined by a log-normal distribution, captures potential energy efficiency gains per query; we set $\propto = 1$ for all baseline energy computations and, in latter sections, estimate the $\propto$ for various scenarios based on literature.

$L_{\text{in}}, L_{\text{out}}$ denote the number of input and output tokens for a query. We model a variety of workloads by sampling $L_{\text{in}}, L_{\text{out}}$, choosing discrete values for $L_{\text{in}}$ and sampling $L_{\text{out}}$ from an exponential distribution. We analyze two regimes: i) *Traditional*, common in conversational queries with no intensive test-time scaling,[16] with median $L_{\text{out}}$=300 tokens (IQR: 129-618) , and ii) *Test-time scaling*, consisting of long queries common in reasoning models such as OpenAI o3[17] or DeepSeek-R1[18] or long agentic workflows, with median $L_{\text{out}}$ = 5,000 (IQR: 2,040-9,717).[19] $L_{\text{eff}}$ denotes the effective token length and scales the energy by the number of tokens in the query. Since the output tokens dominate energy use, we approximate $L_{\text{eff}} \sim L_{\text{out}}$ and fix $L_{\text{in}}$=500 in our main analysis. In the Supplemental Note S3, we explore additional $L_{\text{in}}$ and $L_{\text{eff}}$ configurations. Although our definition of a query is a single prompt–completion event, multi-step and agentic workflows can be modeled as multiple queries per user request, with total energy approximated by the total output tokens generated across those steps (additional orchestration, tool calling and long context management overhead leading to increased energy use are out of scope of this study).

$TPS$ is the token throughput per second. TPS varies according to the model architecture, deployment configuration, concurrency, and latency constraints. In our case, since we are interested in modeling energy usage in a steady-state deployment, we use the existing benchmark data for H100 using NVIDIA's TensorRT-LLM serving framework[20,21] considering parallelism of 8 (10 in DeepSeek-R1) and continuous batching. As presented in Supplemental Note S1, $TPS$ for five models of interest (DeepSeek-R1 671 B, Llama 3.1 405B, Llama 3.1 Nemotron Ultra 253 B, Mixtral 8x22B, and Llama 3.1 70B) has been measured for multiple $L_{\text{in}}, L_{\text{out}}$ configurations. The reported TPS in these benchmarks approximates the throughput of a fully saturated node with high concurrency. For our simulation, $TPS(L_{\text{in}}, L_{\text{out}})$ is a piecewise log-linear regression model trained on this curated data, as detailed in Supplemental Note S2. The log-linear model is trained on available benchmark data of up to 4,000 output tokens. For longer outputs, we hold TPS at the plateau rather than modeling the gradual decline due to growing $KV$ (Key-Value) and attention overhead. This simplification likely underestimates energy for very long generation but provides a conservative baseline. Although TPS and $P_{\text{node}}$ are correlated, for simplicity we treat $TPS$ as independent of $P_{\text{node}}$ in the main analysis and explore a linear dependency in Supplemental Note S4, leading to similar results.

To estimate $E_{\text{query}}$, we sample 10,000 random configurations of $L_{\text{out}}$, $P_{\text{node}}$ and PUE, and compute the corresponding TPS for a given model. To analyze energy implications and efficiency opportunities, we define a "*Baseline"* energy usage scenario by sampling the estimated energy use of three models above 200B parameters. This baseline results approximate the energy consumption of a large LLM deployment with the most common production optimization techniques in modern serving frameworks, including dynamic batching, $KV$ cache, and other common optimizations in inference engines such as vLLM or TensorRT-LLM. However, many hardware, model, and serving-level optimizations are increasingly being adopted in production, such as prefill and generation disaggregation or improved model quantization. To estimate the impact of this line-of-sight energy efficiency opportunities, we compile what we consider the most imminent improvements and estimate their expected gains in throughput or energy per token based on literature. To update our estimates, we modify Eq. 1, defining a multiplier $\alpha$ to scale-up token throughput ($\alpha \cdot$TPS) or energy-per-query ($1/\alpha$). For a fixed compute node, $\alpha$ is modeled as a log-normal distribution with P5-P95 ranges estimated from literature to capture uncertainty of the impact of each potential optimization.

Finally, to aggregate the total energy consumption of a large deployment, we sampled the energy for serving millions of queries in the *"Baseline"* scenario and compute the total energy of serving those queries. As our estimate is performed at the node-level, we scale our total energy estimate by a factor $\beta = 1.31$, accounting for non-uniform demand profile during the day, node redundancy, and networking,[22] as described in Supplemental Note S5.

# 3 RESULTS

## Estimated energy per query for various models

Figure 1 summarizes the energy per query for five dense and mixture-of-expert (MoE) models: DeepSeek-R1 671 B, Llama 3.1 405B, Llama 3.1 Nemotron Ultra 253 B, Mixtral 8x22B, and Llama 3.1 70B. In the traditional regime, Figure 1a, our Monte Carlo estimates align with the measured energy of production-optimized deployments, as presented in Table 1. For example, Llama 3.1 405B is estimated to consume a median of 0.39 Wh per query (IQR: 0.19–0.68), consistent with the ML.ENERGY leaderboard (v3.0)[4] with a median measured energy of 0.21 Wh and with the measurement by Caravaca et al.[23] Similarly, Mixtral 8x22B is estimated to consume a median of 0.06 Wh per query (IQR: 0.03–0.11), close to the ML.ENERGY measurement of 0.11 Wh (v2.0, normalized from BF16 to FP8 as 0.18 Wh/1.7=0.11). Across all production-scale measurements compiled in Table 1, reported values fall within our estimated interquartile range (IQR) with one exception: a single ML.ENERGY measurement for Llama 3.1 70B reporting 0.04 Wh per query, which lies slightly below our IQR. Notably, an independent production measurement for the same model reports 0.05 Wh per query,[23] which falls within our estimated range. We further note that most published measurements report GPU-only energy, whereas our estimates intentionally target full-node energy consumption and include PUE, leading to modest and expected offsets when comparing absolute values.

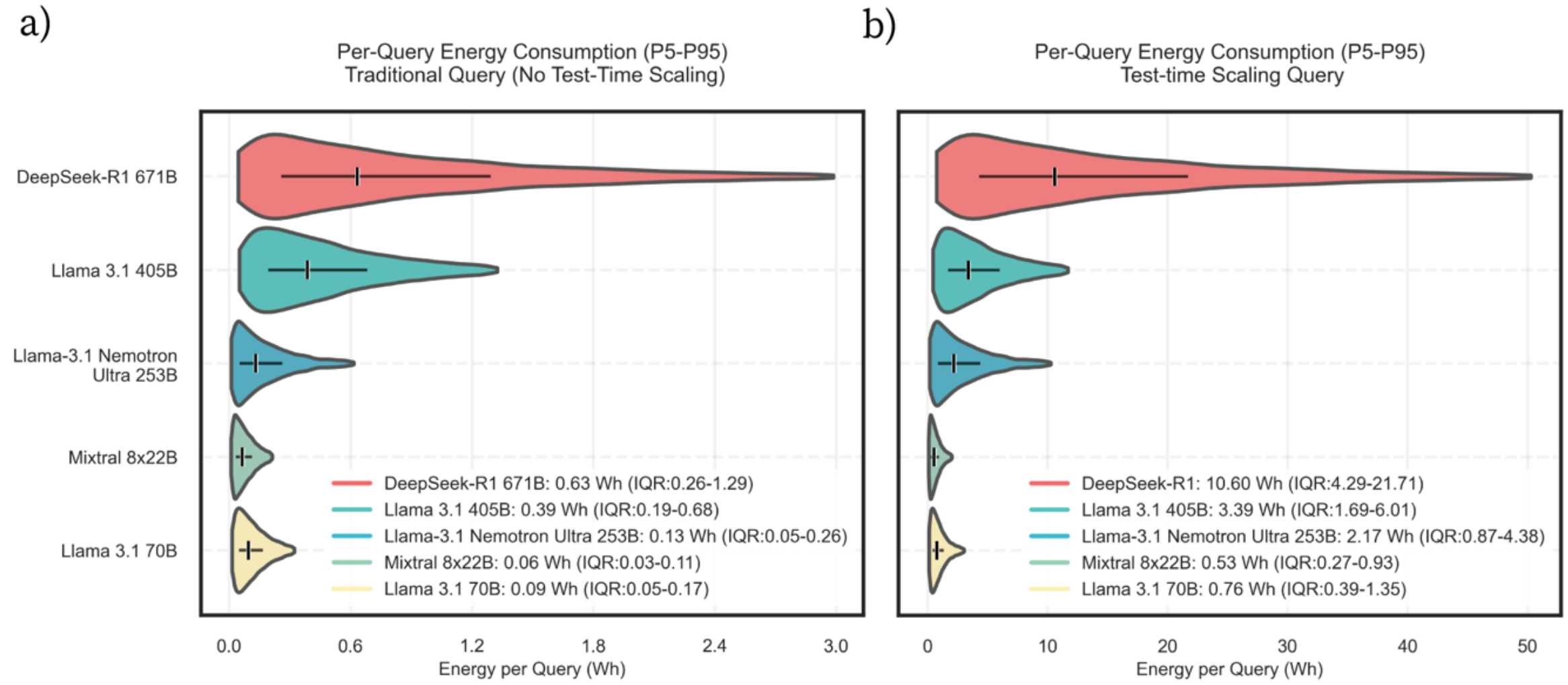

**Figure 1**: **P5-P95 energy per query for several LLMs in two regimes, after sampling 10,000 queries ($L_{out}$, $P_{node}$ and PUE configurations).** a) Standard query, with 300 median output tokens (IQR: 129-618). b) Test-time scaling query, with 5,000 median output tokens (IQR: 2,040-9,717). In both cases, we assume 500 input tokens and estimate token throughput for realistic deployments on an H100 node. Bars represent median and interquartile range.

When in-production deployments are not considered, the energy-per-query is substantially larger: AI Energy Score[6] measured ~1.01 Wh per query for Llama 3.1 70B (average of 1.72 Wh across 1,000 queries with FP16 quantization, ~1.72/1.7=1.01 Wh at FP8), under similar assumptions IEA[24] estimated ~1.25 Wh per-query for Mixtral 8x22B and ~2.25 Wh per-query for DeepSeek-R1. These latter estimates did not consider basic in-production optimization techniques, including batching and inference acceleration, leading to 4–20× overestimation of energy per query. Benchmarking to these numbers or extrapolating them to large-scale inference is not representative of the energy use of major AI products.

Our approach provides an estimate for the energy of test-time scaling queries in Figure 1b. As expected, the per-query energy scales approximately linearly with the total output tokens. These results are consistent with past measurements, as presented in Table 1. DeepSeek-R1 consumes 10.6 Wh (IQR: 4.29–21.7), consistent with the median ML.ENERGY (v3.0) measurement of 20.9 Wh with an average query of 11,287 output tokens. For slightly shorter queries (8,968 average tokens) using the same backbone model as R1 (DeepSeek V3), the median ML.ENERGY measurement was 9.30 Wh. The energy cost of very long queries can be substantial, but those models that maintain high throughput for long queries, such as Mixtral 8x22B, seem to have a scaling advantage at the same number of parameters.

**Table 1:** Reported energy per query (Wh/query) from prior studies compared with our estimates. Reported values derived from production-scale measurements or estimates fall largely within our estimated interquartile range (IQR), while reports based on non-production settings fall outside it. When other hardware or quantization configurations are used and sufficient detail is available, reported values are normalized to H100 and FP8 equivalents, as described in the "Comments" column. For this normalization, we assume a throughput gain of 1.7x for quantization at FP8 compared to BF16 or FP16, corresponding to the average gain reported in by NVIDIA[34], and we assume Blackwell normalized throughput gain of 2.8x compared to H100.[56]

| Source | Type | Model / Claim | Reported Wh/query (H100 and FP8 equivalent when possible) | Regime | Comparable Wh/query In this study | Within estimated IQR? | Comments |
|---|---|---|---|---|---|---|---|
| Altman, 2025 [11] | Disclosure | "Average energy use of | 0.34 | Traditional | >200B Baseline 0.31 | ✅ | No query length, hardware or model mix provided |

| | (Unclear if measured or estimated) | a ChatGPT query” | | | (IQR: 0.16–0.60) | | |
|---|---|---|---|---|---|---|---|
| Elsworth et al., 2025 [2] | Measurement | “Typical Gemini Apps request” | 0.24 | | | ✅ | Fleetwide measurement. No query length, hardware or model mix provided |
| You, 2025 [10] | Estimate | GPT-4o | 0.30 | | | ✅ | Heuristic FLOPs-based estimate. 500 output tokens. |
| Jegham et al., 2025 [8] | Estimate | GPT-4o | 0.42 | | | ✅ | API user-side throughput for 300 output tokens |
| De Vries, 2023 [9] | Estimate | “Average ChatGPT query” | 2.9 | | | ❌ | Outdated estimate. Non-production conditions, no batching, no serving engine. |
| ML.ENERGY Benchmark (version 2.0), 2025 [4] | Measurement | Mixtral 8×22B | 0.11 | Traditional | Mixtral 8x22B 0.06 (IQR: 0.03–0.11) | ✅ | Median energy per query using BF16 reported as 0.18 Wh. Adjusted to FP8 quantization: 0.18 Wh/1.7=0.11 Wh Average output tokens: 389. |
| ML.ENERGY Benchmark (version 3.0), 2025 [4] | Measurement | Llama 3.1 70B | 0.04 | Traditional | Llama 3.1 70B 0.09 (IQR: 0.05–0.17) | ❌ | Median energy per query for 4xH100 using BF16 reported as 0.04 Wh. Assumed to be equivalent to 8xH100 using FP8. Average output tokens: 361. |
| Caravaca, 2025 [23] | Measurement | Llama 3.1 70B | 0.05 | Traditional | | ✅ | Reported for FP16 as 0.08 Wh/query. We approximate to FP8 as 0.08 Wh/1.7=0.05. 300 output tokens. |
| ML.ENERGY Benchmark (version 3.0), 2025 [4] | Measurement | Llama 3.1 405B | 0.21 | Traditional | Llama 3.1 405B 0.39 (IQR: 0.19–0.68) | ✅ | Median energy per query using FP8 reported as 0.21 Wh. Average output tokens: 367. |
| Caravaca, 2025 [23] | Measurement | Llama 3.1 405B | 0.21 | Traditional | | ✅ | Reported for FP16 as 0.35 Wh/query. We approximate to FP8 as 0.35 Wh/1.7=0.21. 300 output tokens. |
| ML.ENERGY Benchmark (version 3.0), 2025 [4] | Measurement | DeepSeek-V3 | 9.30 | Test-time scaling | DeepSeek-R1 10.6 (IQR 4.29–21.7) | ✅ | Median energy per query for B200 using FP8. Adjusted to H100 as 3.32 Wh *2.8=9.30, based on [56]. Average output tokens: 8,968. |
| ML.ENERGY Benchmark (version 3.0), 2025 [4] | Measurement | DeepSeek-R1 | 20.9 | | | ✅ | Median energy per query for B200 using FP8. Adjusted to H100 as 7.45 Wh*2.8=20.9. Average output tokens: 11,287. |
| AI Energy Score, 2025 [6] | Measurement | Llama 3.1 70B | 1.01 | Traditional | Llama 3.1 70B 0.09 (IQR: 0.05–0.17) | ❌ | We adjusted for quantization as 1.72 Wh/1.7=1.01. Non-production conditions, no batching, no serving engine. |
| IEA estimate, 2025 [24] | Estimate | Mixtral 8×22B | ~1.25 | Traditional | Mixtral 8x22B 0.06 (IQR: 0.03–0.11) | ❌ | Adjusted for quantization. Non-production |
| | | | | | | | conditions, no batching, no serving engine. |
| IEA estimate, 2025 [24] | Estimate | DeepSeek-R1 | ~2.25 | Traditional | DeepSeek-R1 0.63 (IQR: 0.19–0.68) | ✗ | Adjusted for quantization. Non-production conditions, no batching, no serving engine. |

## Line-of-sight reductions in energy use

To analyze energy implications and efficiency opportunities, we focus on line-of-sight improvements, defined as energy-efficiency interventions that are already deployed or close to production in some serving configurations and are credibly expected to be adopted in large-scale inference deployments within the next hardware or model generation. We defined a "*Baseline"* energy usage scenario by sampling the estimated energy use of the three models above 200B parameters (Figure 1a: DeepSeek-R1 671B, Llama 3.1 405B, Llama 3.1 Nemotron Ultra 253B). This parameter threshold is based on the size of popular, frontier-scale models, including open weight models like DeepSeek-R1[18], Llama 3.1 405B[25], Qwen 3[26], and is also a reasonable range for proprietary models such as GPT-4o or Claude 3.5 Sonnet.[27] As presented in Figure 2, the median energy per query in a baseline of models over 200B parameters is 0.31 Wh (IQR: 0.16–0.60). As presented in Table 1, this is consistent with recent estimates for a typical chatbot query in a highly-optimized deployment: 0.3 Wh for a typical GPT-4o query of 500 output tokens (heuristic estimate based on FLOPs of a 200B model)[10], 0.42 Wh for GPT-4o query with 300 output tokens (based on user-side token throughput) ,[8] 0.34 Wh for the "average energy use of a ChatGPT query", as disclosed by Sam Altman [11], and 0.24 Wh for "the typical Gemini Apps query".[2] Although not a standardized comparison due to a diversity of products, models, workloads and measurement approaches, this energy range is substantially lower than previous, widely reported estimates.[9,24] For context, it is also in the range of the commonly cited 0.3 Wh per web search reported in a 2009 Google disclosure.[28] We emphasize that this search figure is dated and likely overestimates present-day web search energy; to our knowledge, no newer public web search disclosures are available. Moreover, web search and LLM inference represent fundamentally different workloads. We therefore use this comparison only as a coarse historical reference point, rather than as a direct benchmark. Various factors have led to this improvement in LLM energy per query compared to the previous generation of models: reduced size of new frontier models (shift from Kaplan to Chinchilla scaling laws, leading to smaller models trained with more data),[27] hardware efficiency gains ,[29] and model and serving improvements.

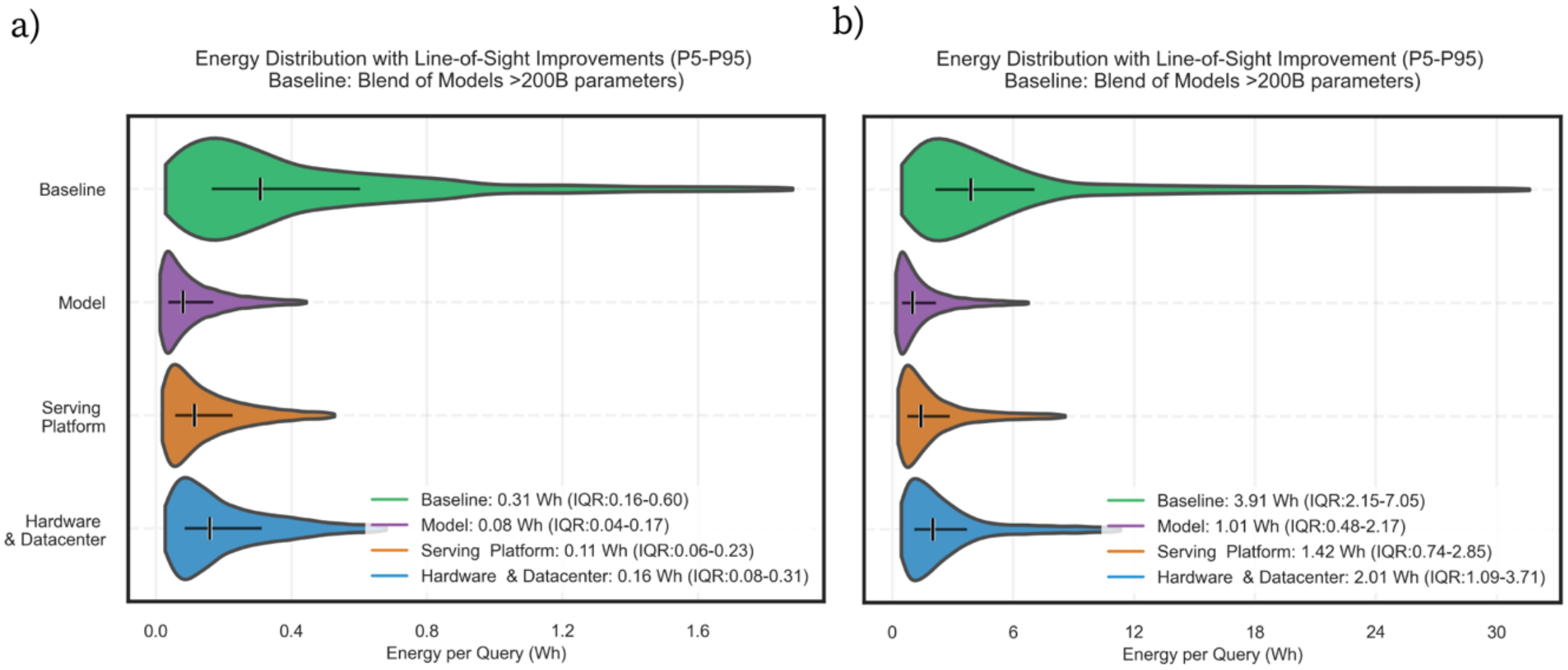

**Figure 2 Estimates of baseline and line-of-sight optimized scenarios for LLMs with over 200B parameters, after sampling 10,000 queries ($L_{out}$, $P_{node}$ and PUE configurations).** P5-P95 energy per query in two regimes: a) Traditional query, with 300 median output tokens (IQR: 129-618). b) Test-time scaling query, with 5,000 median output tokens (IQR: 2,040-9,717). In both cases, we assume 500 input tokens and estimate token throughput for realistic deployments on an H100 node. Each category of energy improvement is considered independent and non-additive compared to the others. Bars represent median and interquartile range.

In the test-time scaling regime (5,000 median output tokens), sampling of models over 200B parameters leads to a median estimate of 3.91 Wh per query (IQR: 2.15–7.05) with over 10 Wh in a significant portion of queries. Given this substantial footprint, even a small portion of test-time scaling queries or token-intensive agentic workflows can have a major impact on the total energy consumption of AI inference. From our perspective, this represents a key frontier for energy efficiency.

For this purpose, we define several of these line-of-sight energy efficiency opportunities and quantify their potential impact. As described in methodology, we defined a multiplier $\alpha$ (P5-P95 range) estimated from literature to capture uncertainty of the impact of each optimization (set as conservative-to-optimistic bounds from reported results and treated as non-additive within each category). We categorize these improvements into:

i) **Model ($\boldsymbol{\alpha = [P5: 1.5, P95: 10]}$):** Algorithm and architecture improvements have been the largest drivers of energy efficiency for inference workloads. Based on recent techniques, we estimate line-of-sight improvements of 1.5–10× for next generation deployments.

- Developments in model distillation have led to energy reductions of 5–10x.[2] Notably, model distillation not only leads to more FLOP-efficient models but also allows larger batch sizes with the same hardware. Distillation has been proven useful for reasoning models: smaller models with longer inference have been shown to outperform larger non-reasoning models, e.g., 7 to 70B reasoning model distilled from DeepSeek-R1 demonstrated comparable performance to models three to five times larger [18], and carefully distilled Qwen 3 reasoning models outperformed larger models.[26] In certain areas, small language models (SLM) can reduce energy by one to two orders of magnitude for token-intensive tasks such as math, coding, or specific agentic workflows compared to models above 100 billion parameters.[30–32] Traditional NLP techniques can also reduce energy use by orders of magnitude compared to LLMs in tasks such as text classification.[33]

- Model quantization already has substantial impact in production: FP8 in H100 already delivers ~1.4–2.4× TPS over FP16.[34] Emerging low-bit quantization techniques supported in newer hardware can plausibly increase inference TPS by 1.5–3× if model performance holds steady.[35,36] Although quantized models could allow efficient parallelism and request management in large deployments, technical challenges for running heavily-quantized models with large batches must be addressed.[37]
- MoE has demonstrated success in reducing inference compute compared to equivalent dense models. Multiple MoE inference optimization approaches are emerging, including optimized expert parallelism (e.g., DeepSpeed-MoE[38]), dynamic gating and load balancing (e.g., Tutel[39]), and expert offloading/caching (e.g., MoE-Lightning[40]). These approaches report 1.3–8× inference speedups depending on baseline and workload. Additionally, attention operations and architectures have been optimized for efficient inference, such as FlashAttention[41] or Multi-Head Latent Attention[18], and long sequence generation.
- Focusing on efficient reasoning models, Llama-Nemotron[21] demonstrated 1.9–4× TPS compared to DeepSeek-R1 using neural architecture search, selective attention removal, and feedforward network optimization. Hybrid attention models using variations of linear or sliding attention have achieved gains for test-time scaling[42,43], e.g., MiniMax-M1[43] consumes 4× fewer FLOPs generating 100K tokens compared to DeepSeek-R1. Other emerging techniques are adaptive computing and controllable reasoning-length generation, allowing models to dynamically reduce the number of generated tokens by up to 40% while maintaining good accuracy.[44]

**ii) Serving platform and workload management ($\boldsymbol{\alpha = [1.5, 5]}$):** Optimizing model deployments to service-level objectives has a significant impact on the energy and carbon footprint of inference.[1,45,46] Furthermore, large-scale AI systems allow efficiency gains by workflow management, for example routing models according to query complexity. We estimate line-of-sight improvements of 1.5–5× due to these interventions.

- Adaptive serving based on service-level objectives demonstrated over 50% reduction in energy consumption.[45] Disaggregated serving allows separation of the prefill and decoding inference stages, and optimizes each stage with different resources, caching, or accelerators,[47,48] leading to 1.4–2.3× TPS improvement.[45] Due to disaggregation, better resource utilization in the decoding stage is especially impactful in long generation and agentic workflows, with TPS gains of 1.4–4× in distributed settings[48,49], especially when combined with dynamic GPU allocation, smart routing, and low-latency communication.[46]
- KV cache management and quantization are critical for efficient model serving; techniques such as LServe, KVQuant, and LMCache enable both long context management and long sequence generation with over 2-15× throughput and inference speedup.[42,50,51] Speculative decoding has demonstrated 2–3× TPS speedups.[52]
- Automated or user-triggered model routing can have a large impact in token throughput and memory and token utilization and thus has become popular in the next-generation of models, such as GPT-5. This is particularly useful for reasoning models: we estimate that models that allow explicit switching of reasoning capabilities can reduce per-query energy by 5x or more due to reduced query length. Smart model routing can actively maintain response performance and reduce per-query energy and cost, e.g., routing with continual learning has been able to maintain 97% response quality with ~2–4x cost reduction.[53,54] Dynamic filtering of low-quality reasoning traces, as done by DeepConf,[55] has demonstrated up to 99.9% accuracy with ~5x query length reduction.

iii) **Hardware and datacenter ($\alpha = [1.5, 2.5]$):** Hardware and datacenter advances are expected to remain a foundation of inference speed-ups. NVIDIA's Blackwell architecture delivers a substantial improvement in TPS per watt (W) compared to H100/H200 GPUs. In MLPerf Inference v5.0 running Llama 3.1 405B, a single Blackwell B200 achieved up to 2.8–3.4× higher TPS per GPU than H100.[56] SemiAnalysis notes that Blackwell improves TFLOPS per GPU-W by 47-82% relative to H100 at the same quantization level.[29] Beyond GPUs, custom AI inference chips (e.g., ASIC, FPGA) have TPS per W gains of around 5–20×,[57] in part due to better performance in the memory-bound decoding stage, a critical capability for test-time scaling. Furthermore, power management techniques such as oversubscription, voltage/frequency scaling and power capping can lead to up to allocating 30% more servers in existing clusters, and 20% peak power reduction.[12] At the datacenter level, cooling optimization can provide additional gains, e.g., cold-plate and immersion cooling can reduce energy demand by 15–20% compared to air cooling.[58] Taken together, without considering improved quantization or custom hardware, these data points support a TPS per GPU-W increase of 1.5–2.5×.

Figure 2 summarizes the estimated impact of these line-of-sight interventions by re-running the Monte Carlo simulation with a category-specific efficiency multiplier $\alpha$ applied to Eq. (1), while holding the other categories at baseline; therefore, these savings are independent of each other. In the traditional regime, hardware/datacenter improvements and serving/workflow optimization reduce the median energy per query from 0.31 Wh to 0.16 Wh and 0.11 Wh, respectively (roughly ~2× and ~3× reductions), while model improvements reduce it further to a median of 0.08 Wh (~4×), conditional on maintaining competitive model performance. In the test-time scaling regime, the relative reductions are similar, and the long-tail (P95) energy consumption is reduced. Model and serving optimizations enable intensive test-time scaling: a reasoning query with a median of 5,000 output tokens can be served at ~1–1.5 Wh, comparable to much smaller queries without optimization. If adopted jointly across layers, these gains can compound multiplicatively. For a conservative illustrative stack (about ~1.5× from hardware/datacenter advances, ~2× from serving/workflow optimization, and ~2.5–3× from model improvements), the overall reduction is ~8–9× (e.g., 1.5×2×2.5≈8×). For a representative optimistic efficiency gain (~2× hardware, ~2.5× serving, ~4× model), the overall reduction is ~20× (2×2.5×4=20×). This combined range reflects realistic stacking across layers rather than the product of category endpoints, which would overstate achievable gains due to overlap and deployment trade-offs; the resulting magnitude is consistent with reductions reported in large-scale production systems (33x).[2]

## Energy use of a billion queries per day

Based on the discussed scenarios, we extrapolate the energy consumption of serving 1 billion queries per day. The billion queries scale aligns with usage patterns reported across leading conversational AI platforms,[59,60] such as ChatGPT at 2.5 billion queries per day as of July, 2025.[60] Queries on dominant web search platforms, in comparison, are estimated at around 10 billion per day.[61] According to our methodology, we sampled the energy required to serve 1 billion queries in the Baseline scenario and aggregated the total energy of serving those queries. We used a scale factor $\beta$=1.31 to account for non-uniform demand, networking, and redundancy at rack and datacenter level. Figure 3a presents the total energy use of 1 billion traditional queries per day. The naïve “Baseline” scenario leads to 0.73 GWh, whereas a very conservative line-of-sight improvement ($\alpha$ = [1.5,3]) results in less than half the total energy use, 0.33 GWh. As reference, a single 30 MW H100 data center operating at constant 80% utilization will consume under 0.6 GWh of energy per day, comparable to the “Baseline” scenario. Using the stated energy use per web search, 1 billion web searches would consume approximately 0.3 GWh. Thus, we observe that our conservative line-of-sight improvement for AI inference reduces the traditional AI inference energy footprint substantially. In contrast to past claims[2,6,24], given that inference can be distributed across geographies and electrical grids, this is a large but manageable load: even if AI inference scaled to 10 billion queries per day to be on par with the magnitude of web search, this would result in energy use of around 1.2 TWh/year. That energy use is about ~5.5% of the current energy use by large hyperscalers such as Microsoft or Google, and ~1% of the current US data center electricity consumption.[24,62] The greatest strain on power grids from new AI workloads is not from inference energy use, but from training loads, the rapid rate of AI adoption, and concentrated capacity build-up –factors which are more challenging to distribute and optimize.[24]

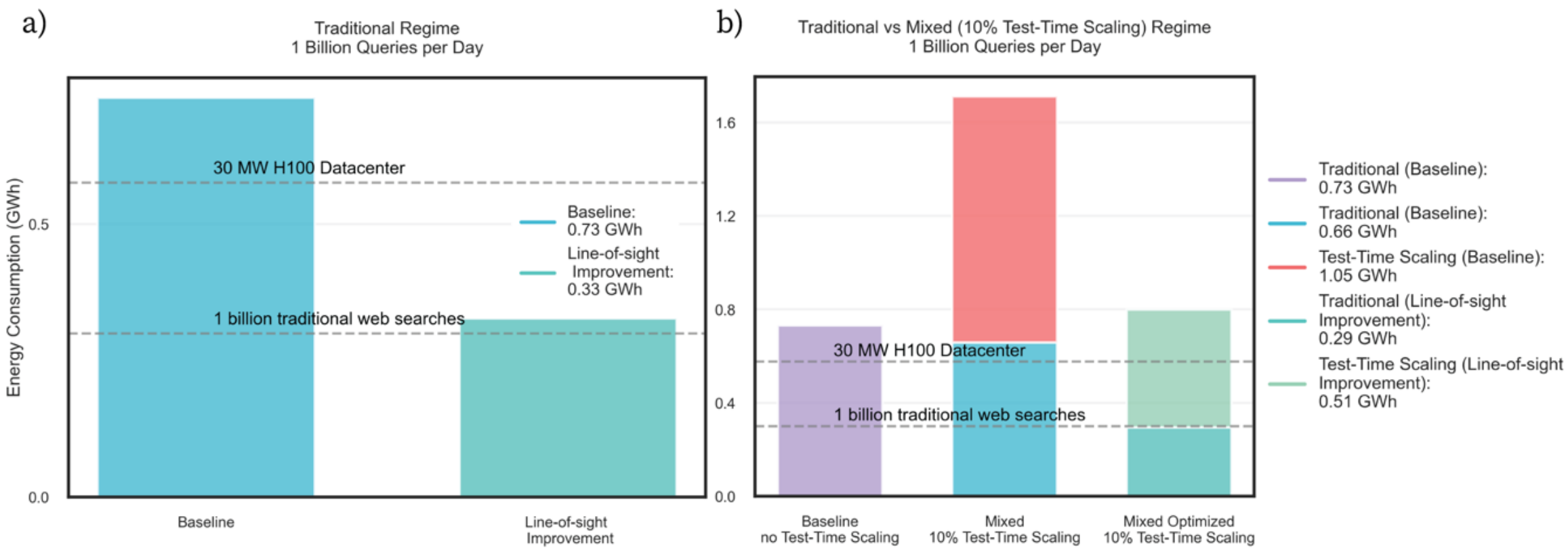

**Figure 3: Energy required for serving 1 billion queries per day.** a) Energy in the traditional regime with a median output length of 300 tokens (IQR: 129-618). Baseline estimate (model > 200B parameters) is compared to a conservative improved scenario with efficiency gains of $\alpha$ =1.5–3. b) Energy in the traditional regime compared to a mixed scenario with 10% test-time scaling queries with median of 5,000 output tokens (IQR: 2,040-9,717). Although very long inference has a large effect on energy, the conservative improved scenario with efficiency gains of 1.5–3x moderates energy growth.

A potential source of concern, however, is the test-time scaling regime. In Figure 3b, we model a mixed inference scenario where 90% of queries are traditional, while 10% are long-inference queries. We chose this 90/10 per-query split to approximate the incidence of long-inference queries in production; given that test-time scaling queries are typically an order of magnitude longer than standard responses, such a mix naturally yields token distributions in which test-time scaling account for roughly half of total tokens served, consistent with public OpenRouter telemetry for 100 trillion tokens [63]. A naïve estimate, not accounting for potential improvements, results in multiplicative energy growth: having 10% of total queries be long inference can more than double the daily energy footprint of serving 1 billion queries in the “Baseline” scenario. However, line-of-sight efficiency improvements can effectively reduce this energy growth. In the optimized scenario with 10% test-time scaling queries, this results in a total energy consumption of 0.8 GWh for 1 billion queries, equivalent to the daily energy consumption required to power roughly 0.4% of television watching in US households.[64]

And although algorithmic advances may produce the largest efficiencies, model routing and workflow management demonstrate impact on energy use even with limited reliance on more complex model optimization. Thus, although test-time scaling can substantially raise energy consumption, its multiplicative effect can be mitigated through line-of-sight optimizations—and because cost per token is the dominant driver of AI adoption, efficiency gains in long inference should be expected and actively pursued.

Historically, projections of digital infrastructure energy use have repeatedly overestimated demand by ignoring rapid gains in efficiency and system-level optimization amid multiplicative demand growth.[62] For instance, a 1999 forecast warned that the internet could consume 50% of the U.S. grid by 2010, but actual data center usage peaked around 2%.[62] Another high-profile projection in 2015 estimated data centers would consume 8,000 TWh/year by 2030—more than 25% of global electricity—but updated estimates now suggest 600–800 TWh, or about 2%.[62] In our opinion, the current surge in AI inference demand may seem unprecedented, but it appears to repeat the same historical pattern: while absolute energy use will grow, efficiency gains in hardware, software, and deployment strategies at scale can moderate its long-term energy footprint.

## 4 CONCLUSION

In this work, we proposed a bottom-up methodology to estimate the energy use of AI inference. By combining node-level power parameters, utilization factors, PUE distributions, and empirically fitted token throughput under realistic serving conditions, our estimates align with disclosures from optimized production environments. This framework enables us to quantify how existing optimization techniques and token-length distributions shape overall energy consumption.

This study has several limitations. First, our simulation focuses on single-node inference with relatively high utilization, typical of large-scale production deployments, and does not fully capture inter-node orchestration overhead, ramp-up dynamics, token throughput and power utilization correlation, or network latency. Second, we assume an idealized token-length distribution for both traditional and long inference, which may differ across enterprise and consumer deployments. Also, agentic workflows could be very heterogeneous, leading to their energy usage being dominated by tool calling or multiple parallel calls rather than single long generation. Third, our modeling of prefill costs in long-context queries is limited;[65] in extreme scenarios (e.g., 100 thousand input tokens), prefill energy use will be substantially higher. This limitation is particularly relevant for agentic coding workloads, multi-document summarization or retrieval workflows;[65] in practice, context management, tool use (e.g., *grep* for code search), disaggregated prefill, and conversation / KV caching[51] significantly reduce this effect. Fourth, our proposed TPS model is chosen as an interpolation model within the decoding-dominant regime, assuming a plateau of token throughput for very long generations, even though real systems often show decline; this likely leads to an underestimate of energy in such cases. Finally, our analysis focuses on autoregressive text generation and does not extend to diffusion-based image or video generation, which have different serving constraints and higher energy utilization.[4,6]

Efficiency gains do not guarantee lower absolute energy use. Lower energy per query, lower cost per token or increases in model size can enable higher usage, longer generations, and more token-intensive workflows. This "rebound effect"[66] can increase total inference energy even as systems become more efficient. Therefore, system-level outcomes depend jointly on efficiency improvements and demand dynamics and should be evaluated together in future studies. At the same time, many tasks do not require the most energy-intensive models, since performance often shows diminishing returns beyond a certain model size or amount of test-time compute,[67] motivating right-sizing via routing and specialization.

Nevertheless, our analysis leads to two central conclusions, each with direct technical and policy implications:

1. **Non-production energy use per query is overestimated.** Energy estimates or measurements based on small-scale or non-optimized deployments overestimate the energy use of inference by 4–20×. For policymakers and stakeholders, this means that alarmist claims drawn from narrow benchmarks should be avoided. Instead, disclosures and estimates should emphasize ranges of energy use that reflect workload and serving diversity, utilization, and deployment scale. Comparisons must be performed with a systems approach, in the context of products, rather than relying on comparing isolated models.
2. **More and longer inference increases energy use, but efficiency gains can limit the increase.** Rapid AI adoption, test-time scaling and agentic workflows substantially raise the inference energy use, but their impact can be managed. The largest efficiency gains are expected from algorithmic advances—particularly due to small/distilled LLMs and inference-optimized architectures—and from intelligent model routing, which can prevent unnecessary use of large models. Hardware improvements and serving optimization also contribute meaningfully. Together, these levers can plausibly yield 8–20× line-of-sight reductions. Realizing these per-query efficiencies promptly, along with managing and increasing demand and changing usage dynamics, will be critical to moderating the energy impact of AI inference as it scales.

## RESOURCE AVAILABILITY

Lead Contact: Felipe Oviedo, felipe.oviedo@microsoft.com
Materials Availability: Non-applicable
Data and Code Availability: Analysis scripts will be open sourced at the time of publication. All data used in this study is of public domain.

## ACKNOWLEDGMENTS

The authors thank William Chappell, Lucas Meyer, Rahul Dodhia and Karin Strauss for useful discussion.

## AUTHOR CONTRIBUTIONS

F.O., A.L., A.K., F.K., E.C., M.N. formulated the study. F.O. performed calculations, defined methodology with F.K., E.C. and R.B.. F.O. wrote the manuscript with feedback from all co-authors. J.L.F. supervised the study.

## DECLARATION OF INTERESTS

All authors are Microsoft employees. All analysis is this perspective is based on open-source models and publicly disclosed algorithms and performance metrics.

## REFERENCES

1. Li, Y., Hu, Z., Choukse, E., Fonseca, R., Suh, G.E., and Gupta, U. (2025). EcoServe: Designing carbon-aware AI inference systems. arXiv. https://doi.org/10.48550/arXiv.2502.05043.

2. Elsworth, C., Huang, K., Patterson, D., Schneider, I., Sedivy, R., Goodman, S., Townsend, B., Ranganathan, P., Dean, J., Vahdat, A., et al. (2025). Measuring the environmental impact of delivering AI at Google Scale. arXiv. https://doi.org/10.48550/arXiv.2508.15734.

3. Gupta, U., Kim, Y.G., Lee, S., Tse, J., Lee, H.-H.S., Wei, G.-Y., Brooks, D., and Wu, C.-J. (2021). Chasing carbon: The elusive environmental footprint of computing. In *Proceedings of the 2021 IEEE International Symposium on High-Performance Computer Architecture (HPCA)*, pp. 854–867. https://doi.org/10.1109/HPCA51647.2021.00076.

4. Chung, J.-W., Ma, J.J., Wu, R., Liu, J., Kweon, O.J., Xia, Y., Wu, Z., and Chowdhury, M. (2025). The ML.ENERGY benchmark: Toward automated inference energy measurement and optimization. In *Advances in Neural Information Processing Systems 38 (NeurIPS 2025)*.

5. Luccioni, A.S., Viguier, S., and Ligozat, A.-L. (2023). Estimating the carbon footprint of BLOOM, a 176B parameter language model. *J. Mach. Learn. Res.* 24, 1–15.

6. AI Energy Score (2025). AI Energy Score. Hugging Face, Salesforce. https://huggingface.github.io/AIEnergyScore/.

7. Samsi, S., Zhao, D., McDonald, J., Li, B., Michaleas, A., Jones, M., Bergeron, W., Kepner, J., Tiwari, D., and Gadepally, V. (2023). From words to watts: Benchmarking the energy costs of large language model inference. In *2023 IEEE High Performance Extreme Computing Conference (HPEC)*, pp. 1–9. https://doi.org/10.1109/HPEC58863.2023.10363447.

8. Jegham, N., Abdelatti, M., Koh, C.Y., Elmoubarki, L., and Hendawi, A. (2025). How hungry is AI? Benchmarking energy, water, and carbon footprint of LLM inference. arXiv. https://doi.org/10.48550/arXiv.2505.09598.

9. de Vries, A. (2023). The growing energy footprint of artificial intelligence. *Joule* 7, 2191–2194. https://doi.org/10.1016/j.joule.2023.09.004.

10. You, J. (2025). How much energy does ChatGPT use? *Epoch AI*. https://epoch.ai/gradient-updates/how-much-energy-does-chatgpt-use.

11. Altman, S. (2025). The gentle singularity. *Sam Altman Blog*. https://blog.samaltman.com/the-gentle-singularity.

12. Patel, P., Choukse, E., Zhang, C., Goiri, Í., Warrier, B., Mahalingam, N., and Bianchini, R. (2024). Characterizing power management opportunities for LLMs in the cloud. In *Proceedings of the 29th ACM International Conference on Architectural Support for Programming Languages and Operating Systems, Volume 3*, pp. 207–222. https://doi.org/10.1145/3620666.3651329.

13. NVIDIA (2026). Introduction to NVIDIA DGX H100/H200 systems. *NVIDIA DGX H100/H200 User Guide*. https://docs.nvidia.com/dgx/dgxh100-user-guide/introduction-to-dgxh100.html.

14. Microsoft (2026). Measuring energy and water efficiency for Microsoft datacenters. https://datacenters.microsoft.com/sustainability/efficiency/.

15. Google (2026). Power usage effectiveness. *Google Data Centers*. https://datacenters.google/efficiency.

16. Chiang, W.-L., Zheng, L., Sheng, Y., Angelopoulos, A.N., Li, T., Li, D., Zhu, B., Zhang, H., Jordan, M.I., Gonzalez, J.E., et al. (2024). Chatbot Arena: An open platform for evaluating LLMs by human preference. In *Proceedings of the 41st International Conference on Machine Learning (ICML 2024)*, pp. 8359–8388.

17. OpenAI (2025). Introducing OpenAI o3 and o4-mini. https://openai.com/index/introducing-o3-and-o4-mini/.

18. DeepSeek-AI, Guo, D., Yang, D., Zhang, H., Song, J., Zhang, R., Xu, R., Zhu, Q., Ma, S., Wang, P., et al. (2025). DeepSeek-R1: Incentivizing reasoning capability in LLMs via reinforcement learning. arXiv. https://doi.org/10.48550/arXiv.2501.12948.

19. Emberson, L. (2025). LLM responses to benchmark questions are getting longer over time. *Epoch AI*. https://epoch.ai/data-insights/output-length.

20. NVIDIA (2025). Overview. *TensorRT-LLM 0.19 Documentation*. https://github.com/NVIDIA/TensorRT-LLM/blob/release/0.19/docs/source/performance/perf-overview.md.

21. Bercovich, A., Levy, I., Golan, I., Dabbah, M., El-Yaniv, R., Puny, O., Galil, I., Moshe, Z., Ronen, T., Nabwani, N., et al. (2025). Llama-Nemotron: Efficient reasoning models. arXiv. https://doi.org/10.48550/arXiv.2505.00949.

22. NVIDIA (2025). Data center design featuring NVIDIA DGX H100 systems. *NVIDIA DGX SuperPOD Documentation*. https://docs.nvidia.com/dgx-superpod/design-guides/dgx-superpod-data-center-design-h100/latest/index.html.

23. Caravaca, F., Cuevas, Á., and Cuevas, R. (2025). From prompts to power: Measuring the energy footprint of LLM inference. arXiv. https://doi.org/10.48550/arXiv.2511.05597.

24. IEA (2025). Energy and AI. IEA, Paris. https://www.iea.org/reports/energy-and-ai.

25. Grattafiori, A., Dubey, A., Jauhri, A., Pandey, A., Kadian, A., Al-Dahle, A., Letman, A., Mathur, A., Schelten, A., Vaughan, A., et al. (2024). The Llama 3 herd of models. arXiv. https://doi.org/10.48550/arXiv.2407.21783.

26. Yang, A., Li, A., Yang, B., Zhang, B., Hui, B., Zheng, B., Yu, B., Gao, C., Huang, C., Lv, C., et al. (2025). Qwen3 technical report. arXiv. https://doi.org/10.48550/arXiv.2505.09388.

27. Erdil, E. (2024). Frontier language models have become much smaller. *Epoch AI*. https://epoch.ai/gradient-updates/frontier-language-models-have-become-much-smaller.

28. Hölzle, U. (2009). Powering a Google search. *Official Google Blog*. https://googleblog.blogspot.com/2009/01/powering-google-search.html.

29. Patel, D., and Nishball, D. (2024). Nvidia Blackwell Performance TCO analysis: B100 vs B200 vs GB200 NVL72. *SemiAnalysis*. https://newsletter.semianalysis.com/p/nvidia-blackwell-perf-tco-analysis.

30. Xu, H., Peng, B., Awadalla, H., Chen, D., Chen, Y.-C., Gao, M., Kim, Y.J., Li, Y., Ren, L., Shen, Y., et al. (2025). Phi-4-Mini-Reasoning: Exploring the limits of small reasoning language models in math. arXiv. https://doi.org/10.48550/arXiv.2504.21233.

31. Liu, R., Gao, J., Zhao, J., Zhang, K., Li, X., Qi, B., Ouyang, W., and Zhou, B. (2025). Can 1B LLM surpass 405B LLM? Rethinking compute-optimal test-time scaling. arXiv. https://doi.org/10.48550/arXiv.2502.06703.

32. Gemma Team, Kamath, A., Ferret, J., Pathak, S., Vieillard, N., Merhej, R., Perrin, S., Matejovicova, T., Ramé, A., Rivière, M., et al. (2025). Gemma 3 technical report. arXiv. https://doi.org/10.48550/arXiv.2503.19786.

33. Zschache, J., and Hartwig, T. (2025). Comparing energy consumption and accuracy in text classification inference. arXiv. https://doi.org/10.48550/arXiv.2508.14170.

34. NVIDIA (2026). FP8 quantization. *TensorRT-LLM Documentation*. https://nvidia.github.io/TensorRT-LLM/performance/performance-tuning-guide/fp8-quantization.html.

35. NVIDIA (2025). Introducing NVFP4 for efficient and accurate low-precision inference. *NVIDIA Technical Blog*. https://developer.nvidia.com/blog/introducing-nvfp4-for-efficient-and-accurate-low-precision-inference/.

36. Liu, Y., Wen, J., Wang, Y., Ye, S., Zhang, L.L., Cao, T., Li, C., and Yang, M. (2024). VPTQ: Extreme low-bit vector post-training quantization for large language models. In *Proceedings of the 2024 Conference on Empirical Methods in Natural Language Processing (EMNLP 2024)*, pp. 8430–8446.

37. Frantar, E., Castro, R.L., Chen, J., Hoefler, T., and Alistarh, D. (2025). MARLIN: Mixed-precision auto-regressive parallel inference on large language models. In *Proceedings of the 30th ACM SIGPLAN Annual Symposium on Principles and Practice of Parallel Programming*, pp. 239–251. https://doi.org/10.1145/3710848.3710871.

38. Rajbhandari, S., Li, C., Yao, Z., Zhang, M., Aminabadi, R.Y., Awan, A.A., Rasley, J., and He, Y. (2022). DeepSpeed-MoE: Advancing mixture-of-experts inference and training to power next-generation AI scale. In *Proceedings of the 39th International Conference on Machine Learning (ICML 2022)*, *Proceedings of Machine Learning Research* 162, 18332–18346.

39. Hwang, C., Cui, W., Xiong, Y., Yang, Z., Liu, Z., Hu, H., Wang, Z., Salas, R., Jose, J., Ram, P., et al. (2023). Tutel: Adaptive mixture-of-experts at scale. *Proceedings of Machine Learning and Systems* 5, 269–287.

40. Cao, S., Liu, S., Griggs, T., Schafhalter, P., Liu, X., Sheng, Y., Gonzalez, J.E., Zaharia, M., and Stoica, I. (2024). MoE-Lightning: High-throughput MoE inference on memory-constrained GPUs. arXiv. https://doi.org/10.48550/arXiv.2411.11217.

41. Shah, J., Bikshandi, G., Zhang, Y., Thakkar, V., Ramani, P., and Dao, T. (2024). FlashAttention-3: Fast and accurate attention with asynchrony and low-precision. In *Advances in Neural Information Processing Systems 37 (NeurIPS 2024)*.

42. Yang, S., Guo, J., Tang, H., Hu, Q., Xiao, G., Tang, J., Lin, Y., Liu, Z., Lu, Y., and Han, S. (2025). LServe: Efficient long-sequence LLM serving with unified sparse attention. *Proceedings of Machine Learning and Systems* 7.

43. MiniMax, Chen, A., Li, A., Gong, B., Jiang, B., Fei, B., Yang, B., Shan, B., Yu, C., Wang, C., et al. (2025). MiniMax-M1: Scaling test-time compute efficiently with lightning attention. arXiv. https://doi.org/10.48550/arXiv.2506.13585.

44. Aggarwal, P., and Welleck, S. (2025). L1: Controlling how long a reasoning model thinks with reinforcement learning. arXiv. https://doi.org/10.48550/arXiv.2503.04697.

45. Stojkovic, J., Zhang, C., Goiri, Í., Torrellas, J., and Choukse, E. (2025). DynamoLLM: Designing LLM inference clusters for performance and energy efficiency. In *2025 IEEE International Symposium on High-Performance Computer Architecture (HPCA 2025)*, pp. 1348–1362. https://doi.org/10.1109/HPCA61900.2025.00102.

46. NVIDIA (2025). NVIDIA Dynamo. https://www.nvidia.com/en-us/ai/dynamo/.

47. Patel, P., Choukse, E., Zhang, C., Shah, A., Goiri, Í., Maleki, S., and Bianchini, R. (2024). Splitwise: Efficient generative LLM inference using phase splitting. In *Proceedings of the 51st Annual International Symposium on Computer Architecture (ISCA 2024)*, pp. 118–132. https://doi.org/10.1109/ISCA59077.2024.00019.

48. Hu, C., Huang, H., Hu, J., Xu, J., Chen, X., Xie, T., Wang, C., Wang, S., Bao, Y., Sun, N., et al. (2024). MemServe: Context caching for disaggregated LLM serving with elastic memory pool. arXiv. https://doi.org/10.48550/arXiv.2406.17565.

49. NVIDIA (2025). Dynamo 0.4 delivers 4x faster performance, SLO-based autoscaling, and real-time observability. *NVIDIA Technical Blog*. https://developer.nvidia.com/blog/dynamo-0-4-delivers-4x-faster-performance-slo-based-autoscaling-and-real-time-observability/.

50. Hooper, C., Kim, S., Mohammadzadeh, H., Mahoney, M.W., Shao, Y.S., Keutzer, K., and Gholami, A. (2024). KVQuant: Towards 10 million context length LLM inference with KV cache quantization. In *Advances in Neural Information Processing Systems 37 (NeurIPS 2024)*.

51. Liu, Y., Cheng, Y., Yao, J., An, Y., Chen, X., Feng, S., Huang, Y., Shen, S., Zhang, R., Du, K., et al. (2025). LMCache: An efficient KV cache layer for enterprise-scale LLM inference. arXiv. https://doi.org/10.48550/arXiv.2510.09665.

52. Chen, C., Borgeaud, S., Irving, G., Lespiau, J.-B., Sifre, L., and Jumper, J. (2023). Accelerating large language model decoding with speculative sampling. arXiv. https://doi.org/10.48550/arXiv.2302.01318.

53. Wang, X., Liu, Y., Cheng, W., Zhao, X., Chen, Z., Yu, W., Fu, Y., and Chen, H. (2025). MixLLM: Dynamic routing in mixed large language models. arXiv. https://doi.org/10.48550/arXiv.2502.18482.

54. Ding, D., Mallick, A., Wang, C., Sim, R., Mukherjee, S., Ruhle, V., Lakshmanan, L.V.S., and Awadallah, A.H. (2024). Hybrid LLM: Cost-efficient and quality-aware query routing. In *The Twelfth International Conference on Learning Representations (ICLR 2024)*.

55. Fu, Y., Wang, X., Tian, Y., and Zhao, J. (2025). Deep think with confidence. arXiv. https://doi.org/10.48550/arXiv.2508.15260.

56. NVIDIA (2026). MLPerf AI benchmarks. https://www.nvidia.com/en-us/data-center/resources/mlperf-benchmarks/.

57. Li, J., Xu, J., Huang, S., Chen, Y., Li, W., Liu, J., Lian, Y., Pan, J., Ding, L., Zhou, H., et al. (2024). Large language model inference acceleration: A comprehensive hardware perspective. arXiv. https://doi.org/10.48550/arXiv.2410.04466.

58. Alissa, H., Nick, T., Raniwala, A., Herranz, A.A., Frost, K., Manousakis, I., Lio, K., Warrier, B., Oruganti, V., DiCaprio, T.J., et al. (2025). Using life cycle assessment to drive innovation for sustainable cool clouds. *Nature* 641, 331–338. https://doi.org/10.1038/s41586-025-08832-3.

59. Paul, K. (2025). Microsoft to spend record $30 billion this quarter as AI investments pay off. *Reuters*. https://www.reuters.com/business/microsoft-spend-record-30-billion-this-quarter-ai-investments-pay-off-2025-07-30/.

60. Allen, M. (2025). Altman plans D.C. push to “democratize” AI economic benefits. *Axios*. https://www.axios.com/2025/07/21/sam-altman-openai-trump-dc-fed.

61. Google (2025). AI, personalization and the future of shopping. https://blog.google/products/ads-commerce/ai-personalization-and-the-future-of-shopping/.

62. Kamiya, G., and Coroamă, V.C. (2025). *Data Centre Energy Use: Critical Review of Models and Results*. 4E Energy Efficient End-use Equipment TCP, EDNA. https://www.iea-4e.org/edna/publications/data-centre-energy-use-critical-review-of-models-and-results/.

63. Aubakirova, M., Atallah, A., Clark, C., Summerville, J., and Midha, A. (2026). State of AI: An empirical 100 trillion token study with OpenRouter. arXiv. https://doi.org/10.48550/arXiv.2601.10088.

64. Luers, A. (2025). Net zero needs AI — five actions to realize its promise. *Nature* 644, 871–873. https://doi.org/10.1038/d41586-025-02641-4.

65. Yen, H., Gao, T., Hou, M., Ding, K., Fleischer, D., Izsak, P., Wasserblat, M., and Chen, D. (2024). HELMET: How to evaluate long-context language models effectively and thoroughly. arXiv. https://doi.org/10.48550/arXiv.2410.02694.

66. Luccioni, A.S., Strubell, E., and Crawford, K. (2025). From efficiency gains to rebound effects: The problem of Jevons' paradox in AI's polarized environmental debate. In *Proceedings of the 2025 ACM Conference on Fairness, Accountability, and Transparency*, pp. 76–88. https://doi.org/10.1145/3715275.3732007.

67. Jin, Y., Wei, G.-Y., and Brooks, D. (2025). The energy cost of reasoning: Analyzing energy usage in LLMs with test-time compute. arXiv. https://doi.org/10.48550/arXiv.2505.14733.

# SUPPLEMENTAL

### *Supplemental Note S1. TPS* Benchmark

*Table S 1: Tokens per Second (TPS) in high concurrency settings for open weight models running on H100 nodes.*

| Model | TP Size | Quantization | Tokens per Second (TPS) | Input Length | Output Length | Source |
|---|---|---|---|---|---|---|
| Llama 3.1 405B | 8 | FP8 | 2050.00 | 500 | 2000 | llamaNemotron Article[1] |
| Llama 3.1 405B | 8 | FP8 | 480.00 | 5000 | 500 | |
| Llama 3.1 405B | 8 | FP8 | 3732.40 | 128 | 128 | tensorRT-LLM-May2025 v019[2] |
| Llama 3.1 405B | 8 | FP8 | 4572.23 | 128 | 2048 | |
| Llama 3.1 405B | 8 | FP8 | 2911.42 | 128 | 4096 | |
| Llama 3.1 405B | 8 | FP8 | 3661.85 | 500 | 2000 | |
| Llama 3.1 405B | 8 | FP8 | 2963.36 | 1000 | 1000 | |
| Llama 3.1 405B | 8 | FP8 | 3253.17 | 1000 | 2000 | |
| Llama 3.1 405B | 8 | FP8 | 3089.16 | 1024 | 2048 | |
| Llama 3.1 405B | 8 | FP8 | 448.89 | 2048 | 128 | |
| Llama 3.1 405B | 8 | FP8 | 2139.94 | 2048 | 2048 | |
| Llama 3.1 405B | 8 | FP8 | 579.14 | 5000 | 500 | |
| Llama 3.1 405B | 8 | FP8 | 370.26 | 20000 | 2000 | |
| Llama 3.1 70B | 8 | FP8 | 14686.01 | 128 | 128 | |
| Llama 3.1 70B | 8 | FP8 | 17463.99 | 128 | 2048 | |
| Llama 3.1 70B | 8 | FP8 | 12598.55 | 128 | 4096 | |
| Llama 3.1 70B | 8 | FP8 | 14630.41 | 500 | 2000 | |
| Llama 3.1 70B | 8 | FP8 | 11082.48 | 1000 | 1000 | |
| Llama 3.1 70B | 8 | FP8 | 11108.11 | 1000 | 2000 | |
| Llama 3.1 70B | 8 | FP8 | 11028.32 | 1024 | 2048 | |
| Llama 3.1 70B | 8 | FP8 | 1840.12 | 2048 | 128 | |
| Llama 3.1 70B | 8 | FP8 | 8772.88 | 2048 | 2048 | |
| Llama 3.1 70B | 8 | FP8 | 2399.78 | 5000 | 500 | |
| Llama 3.1 70B | 8 | FP8 | 1568.84 | 20000 | 2000 | |
| Mixtral 8x22B | 8 | FP8 | 21876.08 | 128 | 128 | tensorRT-LLM-Feb2025 v017[3] |
| Mixtral 8x22B | 8 | FP8 | 25150.03 | 128 | 2048 | |
| Mixtral 8x22B | 8 | FP8 | 18387.40 | 128 | 4096 | |
| Mixtral 8x22B | 8 | FP8 | 21421.86 | 500 | 2000 | |
| Mixtral 8x22B | 8 | FP8 | 16573.24 | 1000 | 1000 | |
| Mixtral 8x22B | 8 | FP8 | 2794.97 | 20480 | 128 | |
| Mixtral 8x22B | 8 | FP8 | 12244.93 | 2048 | 2048 | |
| Mixtral 8x22B | 8 | FP8 | 3645.27 | 5000 | 500 | |
| Mixtral 8x22B | 8 | FP8 | 2227.63 | 20000 | 2000 | |
| DeepSeek-R1 | 10 | FP8 | 886.00 | 1024 | 1024 | tensorRT-LLM-DeepSeekR1[2] |
| Llama-3.1 Nemotron Ultra 253B | 8 | FP8 | 5200.00 | 500 | 2000 | llamaNemotron Article[1] |
| Llama-3.1 Nemotron Ultra 253B | 8 | FP8 | 720.00 | 5000 | 500 | |
| Llama-3.1 Nemotron Ultra 253B | 8 | FP8 | 7530.00 | 500 | 128 | |
| DeepSeek-R1 | 10 | FP8 | 1300.00 | 500 | 2000 | |
| DeepSeek-R1 | 10 | FP8 | 378.00 | 5000 | 500 | |

***Supplemental Note S2. TPS*** **regression**

To estimate tokens-per-second (TPS) as a function of input and output lengths, we used a piecewise regression approach. We fit a regression of the form:

$$\log(\text{TPS}) = \beta_0 + \beta_1 \log(L_{\text{in}}) + \beta_2 \log(L_{\text{out}})$$

where $L_{\text{in}}$ is the input token length and $L_{\text{out}}$ is the output token length. We chose a log-linear function, using the natural logarithm in all cases, as it provided a better fit than linear regression.

Predictions are transformed back to the original scale as:

$$\widehat{\text{TPS}}(L_{\text{in}}, L_{\text{out}}) = e^{(\beta_0 + \beta_1 \log L_{\text{in}} + \beta_2 \log L_{\text{out}})}$$

Predictions were capped at the maximum observed $TPS_{max}$ as $L_{\text{out}}$ increases for a fixed $L_{\text{in}}$:

$$\widehat{\text{TPS}} = \min(\widehat{\text{TPS}}, TPS_{max})$$

This capping captures the plateau observed in the benchmarked range but does not model the throughput decline seen in very long generations, so our framework likely underestimates energy use in such cases. Furthermore, we are interested in using this model for interpolation within a decoding-dominant regime, rather than a heavy prefill regime or for extrapolation purposes.

In our LLMs of interest, using this TPS regression model, in-distribution $R^2$ has a median of 0.921 across models, as presented in the table below (Llama-3.1 Nemotron Ultra 253B using directly the maximum TPS as it is reported in decoding-dominant regime).

*Table S 2: $R^2$ for TPS regression for several open weight models*

| Model | Leave-one-out cross-validation $R^2$ |
|---|---|
| DeepSeek-R1 | 0.912 |
| Llama 3.1 405B | 0.941 |
| Llama-3.1 Nemotron Ultra 253B | 0.893 |
| Llama 3.1 70B | 0.932 |
| Mixtral 8x22B | 0.921 |

**Supplemental Note S3. Effect of query input length on energy use**

In these sections, we explore the effect increasing the input length from $Lin = 500$ to $Lin = 1000$ and $Lin = 5000$ for all queries and sampling from an exponential distribution $L_{\text{out}}$. Instead of assuming $L_{\text{eff}} \sim L_{\text{out}}$, we define $L_{\text{eff}} = L_{\text{in}} + L_{\text{out}}$.

<u>Traditional regime:</u>

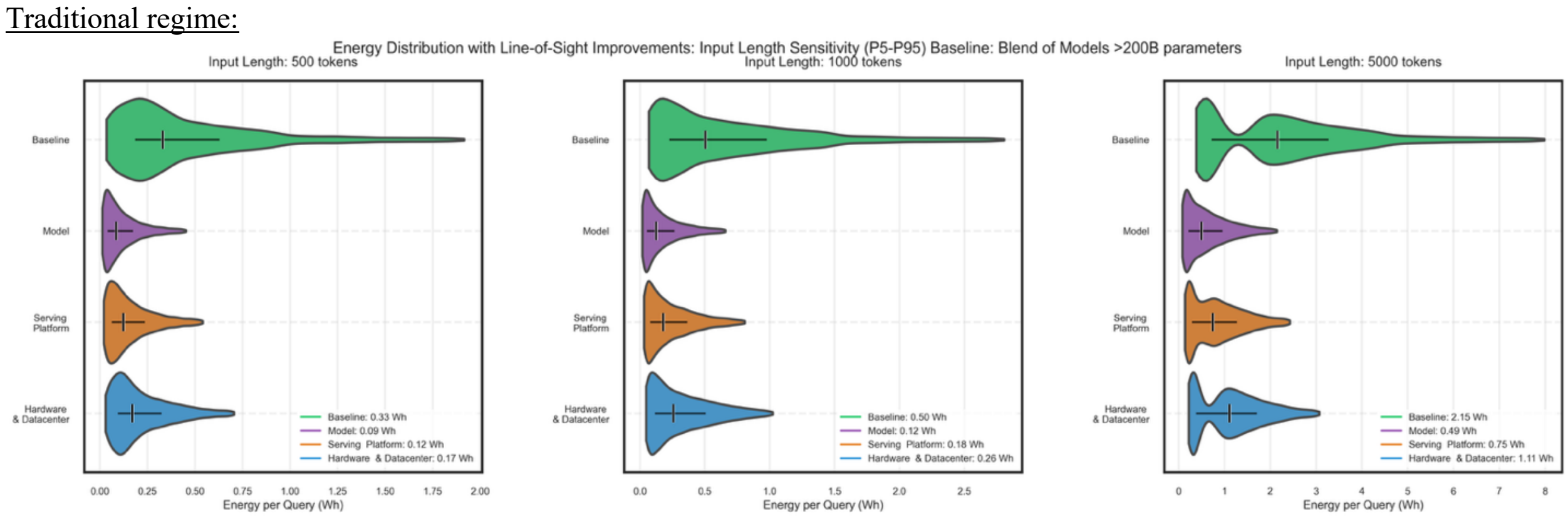

*Figure S 1: Energy per Query (Wh) by token input length in the traditional regime.*

When setting $L_{\text{in}}$ = 500, estimate energy is very close (within 6.5% of median energy per query) to our original Baseline in the traditional regime, which assumed $L_{\text{eff}} = L_{\text{out}}$. $L_{\text{in}}$ = 1000 increases the energy per query, but with a lower effect than increasing $L$out. When $L_{\text{in}}$ = 5000, the energy distribution becomes bimodal because long prefill dominates queries with short outputs.

<u>Test-time scaling regime:</u>

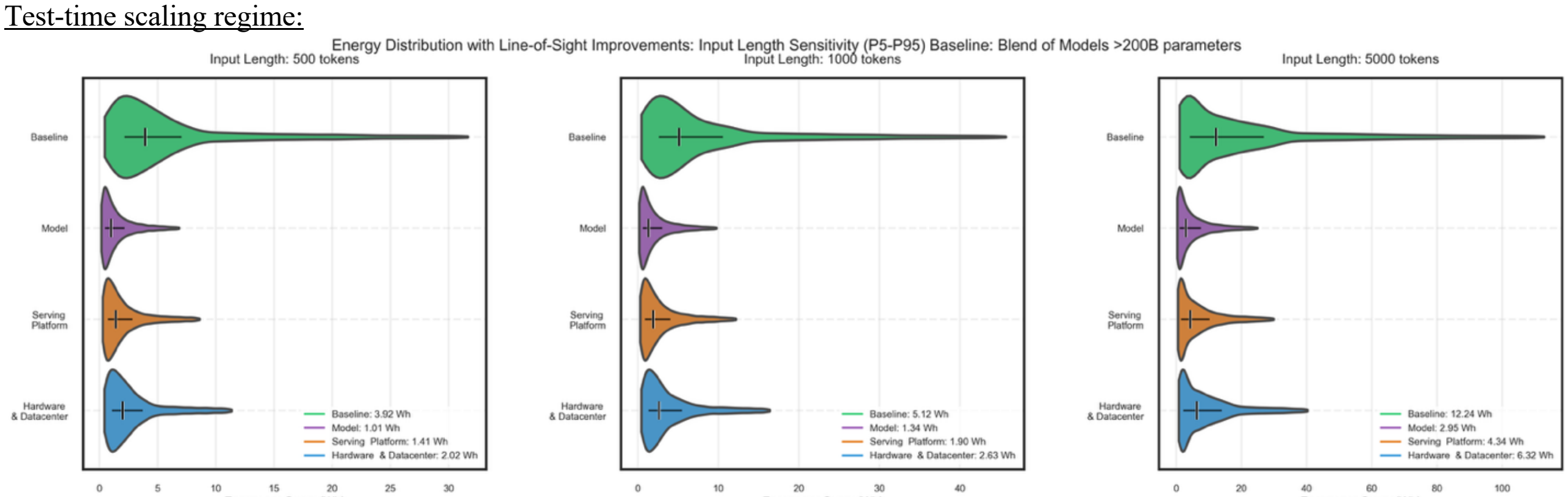

*Figure S 2: Energy per Query (Wh) by token input length in the test-time scaling regime.*

When setting $L_{\text{in}}$= 500, estimate energy is very close (within 0.2% of median energy per query) to our original Baseline in the traditional regime, which assumed $L_{\text{eff}} = L_{\text{out}}$. $L_{\text{in}}$ = 1000 increases the energy per query, but with a lower effect than increasing $L$out. When $L$in = 5000, the energy distribution increases but it does not become bimodal as in the traditional case, as prefill is not dominant.

**Supplemental Note 4. Effect of $TPS$ and $P_{node}$ proportionality**

We explore the effect of increasing $P_{node}$ with increasing $TPS$. For this we assume $P_{node}$ increases linearly with TPS starting from the idle power $P_{idle} = 2.7\ kW$ to $0.9 \cdot P_{max}$

The following figure summarizes the effect on energy per query distributions. For the traditional regime (top) (a) in the baseline scenario, the median energy per query increased only slightly compared to the original analysis with steady-state power as an independent variable (3% increase). For the test-time scaling regime (bottom) (b) the median energy per query increased by 20%, however this is a small effect compared to the effect of longer inference and model size.

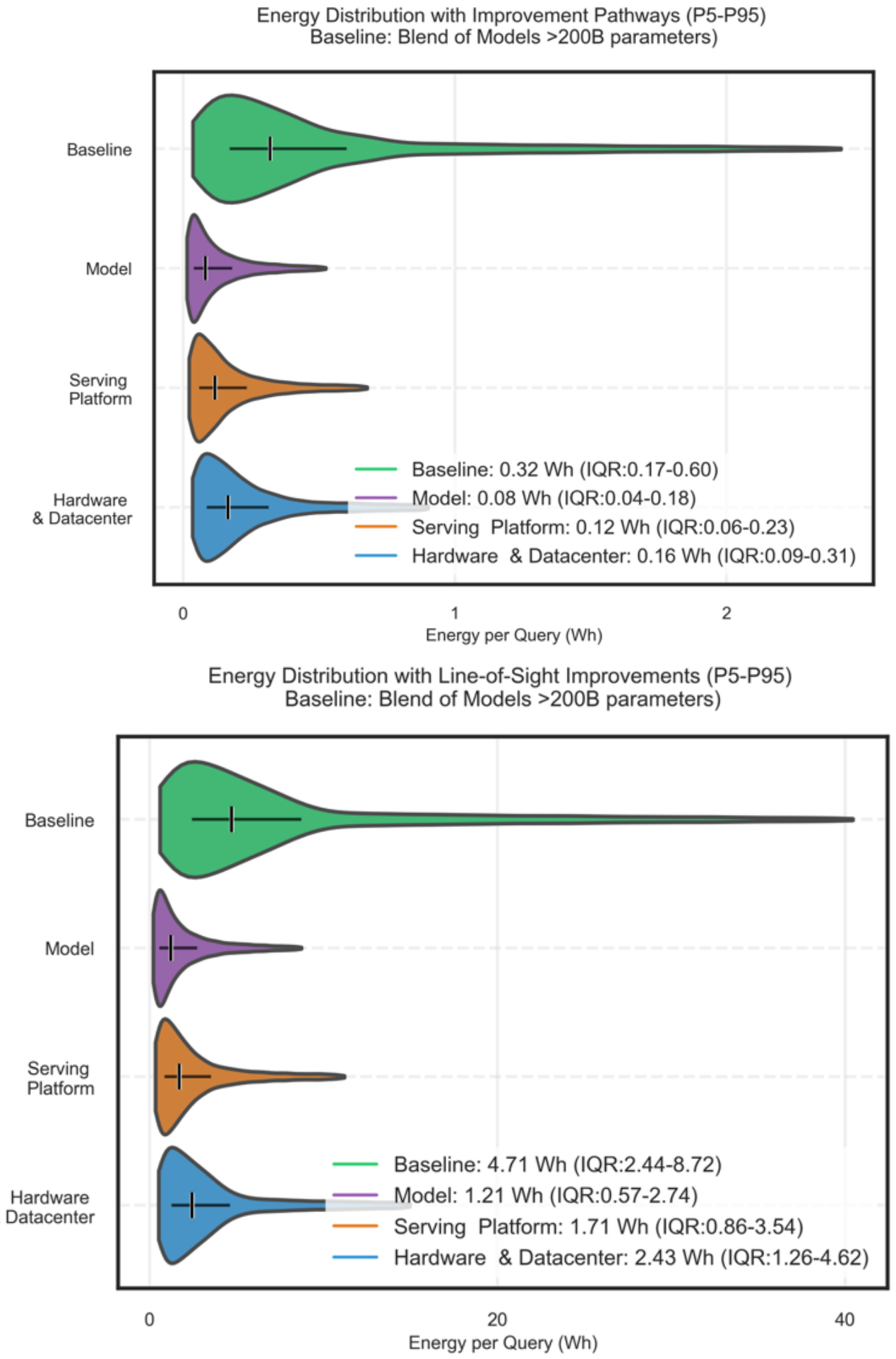

*Figure S 3: Energy per Query (Wh) for traditional and test-time scaling regimes modelling explicitly power and token throughput proportionality.*

**Supplemental Note S5. Estimating scaling factor $\beta = 1.31$**

We estimate the additional number of idle nodes based on a non-uniform usage profile. For this, we assume the daily sinusoidal load with range 50–100%, average 75%. We assume node peak power $P_{max}$ = 11.3 kW, idle power $P_{idle} = 2.7\ kW$, leading to dynamic power $P_{dyn} = 8.6\ kW$

Each node at full load for 24 hours:

$$E_{\text{node}}^{\text{uniform}} = P_{max} \times 24 = 11.3 \times 24 = 271.2\ \text{kWh/day}$$

Each node with sinusoidal load (mean utilization 0.75) consumes per day:

$$E_{\text{node}}^{\sin} = P_{\text{idle}} \times 24 + P_{\text{dyn}} \times 24 \times \overline{f} = 219.6 \text{ kWh/day}$$

So, the scaling factor for utilization is:

$$\beta = \frac{1}{\overline{f}} \times \frac{E_{\text{node}}^{\sin}}{E_{\text{node}}^{\text{uniform}}} = \frac{1}{0.75} \times \frac{219.6}{271.2} = 1.08$$

This estimate corresponds to the excess energy required to serve a variable load between 50% and 100% during the day.

We additionally account for two other factors: (1) network fabric energy and (2) node redundancy. For networking, NVIDIA's DGX H100 SuperPOD[4] reference power breakdown (Table 4 in document) attributes ~8.15% of peak IT power to switching (InfiniBand fabric compute + fabric storage + in-band and out-of-band management switching: 5.72% + 1.91% + 0.45% + 0.07%). We conservatively round this to 10% to capture deployment variability and ancillary fabric overhead.

We assume an additional 10% for node redundancy, based on idle provisioned capacity for reliability/traffic spikes, which is consistent with the fleetwide 8% reported value in [2].

Thus, we have:

$$\beta = 1.10 \times 1.08 \times 1.10 = 1.31$$

## References

**1.** Bercovich, A., Levy, I., Golan, I., Dabbah, M., El-Yaniv, R., Puny, O., Galil, I., Moshe, Z., Ronen, T., Nabwani, N., et al. (2025). Llama-Nemotron: Efficient reasoning models. *arXiv*. https://doi.org/10.48550/arXiv.2505.00949.

**2.** NVIDIA (2025). Overview. *TensorRT-LLM 0.19 Documentation*. https://github.com/NVIDIA/TensorRT-LLM/blob/release/0.19/docs/source/performance/perf-overview.md.

**3.** NVIDIA (2025). Overview. *TensorRT-LLM 0.17 Documentation*. https://github.com/NVIDIA/TensorRT-LLM/blob/release/0.17/docs/source/performance/perf-overview.md.

**4.** NVIDIA (2026). Data center design featuring NVIDIA DGX H100 systems. *NVIDIA DGX SuperPOD Documentation*. https://docs.nvidia.com/dgx-superpod/design-guides/dgx-superpod-data-center-design-h100/latest/index.html.